%% file: main.tex
\documentclass[acmsmall]{acmart}
\AtBeginDocument{%
  \providecommand\BibTeX{{%
    \normalfont B\kern-0.5em{\scshape i\kern-0.25em b}\kern-0.8em\TeX}}}

\setcopyright{acmcopyright}
\copyrightyear{2021}
\acmDOI{10.1145/1122445.1122456}

\acmJournal{TALLIP}

\usepackage{times}
\usepackage{latexsym}
\usepackage{amsmath}
\usepackage{booktabs}
\usepackage{graphicx}
\usepackage[export]{adjustbox}
\usepackage{scalerel}
\usepackage{longtable}
\usepackage{subcaption}
\usepackage{xcolor}
\usepackage[normalem]{ulem}
\usepackage{soul}
\usepackage{color}
\usepackage{tabularx}

\usepackage{multirow}
\usepackage[normalem]{ulem}
\useunder{\uline}{\ul}{}

\usepackage{algorithm}
\usepackage{algpseudocode}

\usepackage[T1]{fontenc}

\usepackage[utf8]{inputenc}

\usepackage{microtype}
\usepackage{soul}

\usepackage{hyperref}

\title{A Constraint-Enforcing Reward for Adversarial Attacks on Text Classifiers}

\author{Tom Roth}
 \orcid{0000-0002-4445-7171}
 \affiliation{%
   \institution{University of Technology Sydney}
   \state{NSW}
   \country{Australia}}
 \affiliation{%
   \institution{CSIRO's Data61}
   \city{Sydney}
   \state{NSW}
   \country{Australia}}
 \email{thomas.p.roth@student.uts.edu.au}
 \author{Inigo Jauregi Unanue}
 \orcid{0000-0001-6223-9584}
 \affiliation{%
   \institution{University of Technology Sydney}
   \state{NSW}
   \country{Australia}}
 \affiliation{%
   \institution{Rozetta Technology}
   \state{NSW}
   \country{Australia}}

 \author{Alsharif Abuadbba}
 \orcid{0000-0001-9695-7947}
 \affiliation{%
   \institution{CSIRO's Data61}
   \state{NSW}
   \country{Australia}}

\author{Massimo Piccardi}
 \orcid{0000-0001-9250-6604}
 \affiliation{%
   \institution{University of Technology Sydney}
   \state{NSW}
   \country{Australia}}

\begin{document}

\begin{abstract}
Text classifiers are vulnerable to adversarial examples --- correctly-classified examples that are deliberately transformed to be misclassified while satisfying acceptability constraints. The conventional approach to finding adversarial examples is to define and solve a combinatorial optimisation problem over a space of allowable transformations. While effective, this approach is slow and limited by the choice of transformations.
An alternate approach is to directly generate adversarial examples by fine-tuning a pre-trained language model, as is commonly done for other text-to-text tasks. This approach promises to be much quicker and more expressive, but is relatively unexplored. For this reason, in this work we train an encoder-decoder paraphrase model to generate a diverse range of adversarial examples. For training, we adopt a reinforcement learning algorithm and propose a constraint-enforcing reward that promotes the generation of valid adversarial examples. Experimental results over two text classification datasets show that our model has achieved a higher success rate than the original paraphrase model, and overall has proved more effective than other competitive attacks. Finally, we show how key design choices impact the generated examples and discuss the strengths and weaknesses of the proposed approach.



\end{abstract}

\ccsdesc[500]{Computing methodologies~Natural language generation}
\ccsdesc[500]{Theory of computation~Adversarial learning}
\ccsdesc[100]{Security and privacy~Software security engineering}
s

\keywords{Adversarial attacks, text generation, natural language processing, reinforcement learning}

\maketitle

\section{Introduction}
\label{sec:introduction}
Adversarial attacks cause a \textit{victim model} --- an attacked machine learning model --- to make a specific mistake.  These attacks occur across domains, pose a real-world security threat\footnote{For example, \citep{wallace2020imitation} attacked Google Translate with adversarial examples, causing vulgar outputs, word flips, and dropped sentences.} and are increasingly well-studied \citep{biggio2018, WeiEmmaZhangSurvey}. In this paper we study adversarial attacks on text classifiers, where an adversary takes a correctly-classified \textit{original} example and perturbs it to create an incorrectly-classified \textit{adversarial} example. The adversarial example must typically meet some acceptability constraints (e.g., a maximum edit distance from the original, preserving semantic meaning, grammaticality), although there is no general consensus on these \citep{textattack}.






\begin{figure} [!ht]
  \centering
    \includegraphics[width=0.8\columnwidth]{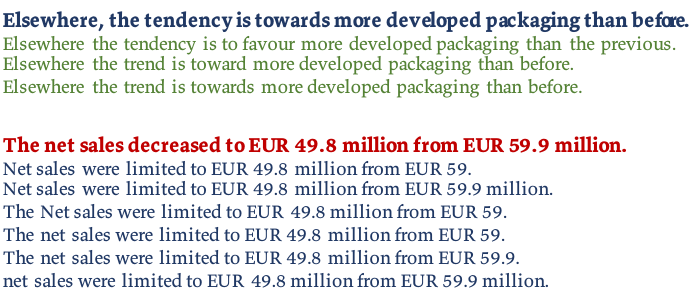}
  \caption{Examples of successful adversarial attacks against a sentiment classifier obtained with the proposed approach. On top, the adversarial examples flip the sentiment from the original neutral (blue) to positive (green), and on bottom, sentiment goes from the original negative (red) to neutral (blue).}
  \label{fig:examples}
\end{figure}

How are text adversarial examples found? The predominant approach is to repeatedly modify tokens until the predicted label changes \citep{WeiEmmaZhangSurvey}. Attacks taking this approach, known as token-modification attacks \citep{roth2021}, find adversarial examples by solving a constrained combinatorial optimisation problem. First, they define the success condition, the constraints and the allowed transformations, and then they use a search algorithm to seek a solution \citep{morris2020reevaluating}. 
While effective, these attacks also have major downsides. Firstly, they are slow: the computational budget heavily impacts their success rate, with high-performing search algorithms requiring many victim model queries per example, particularly for long texts \citep{yoo2020searching}. Secondly, their allowed token-level transformations limit their search space, largely preventing complex transformations like paraphrasing or style change. 



Alternatively, the adversarial example task can be formulated as a text-to-text problem, with original examples as input and adversarial examples as output. It could then be straightforwardly approached with seq2seq models, as done for other text-to-text tasks like summarisation or translation. This approach enjoys several principled advantages over token-modification attacks. Firstly, once trained, finding adversarial examples is much faster (in the order of a few milliseconds, rather than minutes or even hours). Additionally, through beam search or sampling, this approach can easily generate multiple adversarial examples per given input, while also controlling their diversity, tonality, or other characteristics. Finally, a seq2seq approach is also intrinsically more flexible as it is not limited by a rigid set of allowed transformations.

On the other hand, the main challenge of this approach is that it is notoriously difficult to train a model to generate controlled text. The training process can be unstable and time-consuming, and the generated text can be ungrammatical, irrelevant, nonsensical, unnatural, bland, repetitive, or incoherent \citep{Holtzman2020The, hu2017controlledgeneration, dancin_seq2seq}. For our task there is an additional challenge: the generated text must change the victim model's predicted label while not violating any constraint. 


For these reasons, in this paper we propose fine-tuning a pre-trained encoder-decoder paraphrase model so that it produces adversarial examples instead of paraphrases. We fine-tune using a reinforcement learning (RL) policy-gradient algorithm --- REINFORCE with baseline \citep{williams92reinforce} --- and attack a sentiment classifier. For training, we propose an original reward function that both incentivises adversarial examples and penalises any violation of the constraints. To improve generated text coherence, our loss function includes a Kullback-Leibler (KL) divergence term \citep{kldiv}  that limits parameter drift from the pre-trained paraphrase model. The attack requires the victim model's prediction confidence, but no other information, which makes our attack either a \textit{grey-box} \citep{biggio2018} or a \textit{black-box} attack \citep{WeiEmmaZhangSurvey}, depending on the definition.\footnote{These assumptions are not unrealistic: for example, most pre-trained models on the Hugging Face Model Hub report both predictions and confidences.}

We have evaluated the proposed approach on two sentiment analysis datasets, reporting the attack success rates and the diversity of the generated adversarial examples across four different decoding methods and two training temperatures. The results show that the the proposed approach has been able to generate numerous and diverse adversarial examples, with success rates much higher than for the pre-trained paraphraser and comparable token-modification attacks.
In summary, this paper makes the following key contributions:




\begin{enumerate}
    \item an approach for the generation of adversarial attacks to text classifiers based on a pre-trained paraphraser and reinforcement learning;
    \item a constraint-enforcing reward function that incentivises adversarial examples and penalises constraint violations;
    \item experimental results on two text classification datasets showing the effectiveness of the proposed approach, and a comprehensive analysis and discussion. 
\end{enumerate}



 


\section{Related Work}
\label{sec:related}
For ease of reference, we can divide the literature on text classification adversarial attacks into token-modification attacks and generative attacks. 

\textbf{Token-modification attacks.} The vast majority of existing text adversarial attacks are token-modification attacks.
They consist of four main components: a goal function, a set of allowed transformations, a set of constraints that must be satisfied, and a search method \citep{morris2020reevaluating}. These approaches typically create adversarial perturbations by applying repeated token-wise transformations, such as character replacements \citep{HotFlip} or synonym swaps \citep{Ren2019PWWS}\footnote{These approaches typically produce one adversarial example per original. TextAttack \citep{textattack} --- the most popular library for token-modification attacks --- has been set up to only return at most one adversarial example per original, so this is what we have used throughout this paper. While more could be searched for, it has not been well studied how they could be found efficiently in incremental time.}. A detailed description of these attacks is not relevant to our work, so
the reader can refer to a recent survey \citep{roth2021} for further details. 


\textbf{Generative attacks.} 
Some previous work has attempted to train a variety of generative models to produce adversarial examples. For example, long short-term memory variants have been used by \citet{SCPN} to paraphrase a sentence in the form of a parse template, and by \citet{vijayaraghavan2019generating} to perturb examples. A feed-forward network was used by \citet{lu22_good} to generate distracting answers in a multiple-choice visual question answering task. Other work has attempted to use GANs and autoencoders \citep{zhao2018generating, ren202generating,dancin_seq2seq}. However, this line of approach has not been widely pursued, probably due to training difficulties. For example, \citet{dancin_seq2seq} reported widespread issues like mode degeneracy, semantic divergence, and reward hacking. Furthermore, it is challenging to maintain certain text properties, such as label invariance, while generating text.

Since their introduction, transformers have become a ubiquitous encoder-decoder architecture in contemporary natural language processing. They are typically trained with transfer learning, first solving a large-scale pre-training task (typically unsupervised or self-supervised), and then fine-tuning on the target task. Large transformers such as T5 \citep{T5} currently achieve state-of-the-art performance in many text-to-text tasks. Despite this success, no previous work we are aware of has attempted to fine-tune a pre-trained paraphrase model for adversarial example generation, as the proposed approach does. The closest works are \citet{gan2019}, who create a dataset of adversarial paraphrases manually, and \citet{mindthestyle}, who use a pre-trained text style transfer model, but do not fine-tune it.





\section{Proposed Approach}
\label{sec:proposed_approach}

\subsection{Overview}
\label{sec:overview}

Our overall goal is to fine-tune a pre-trained paraphrase model with a reinforcement learning objective so that it can learn to generate adversarial examples. We use a T5 transformer as the base pre-trained model.  

\begin{figure*} [!ht]
  \centering
    \includegraphics[width=\linewidth]{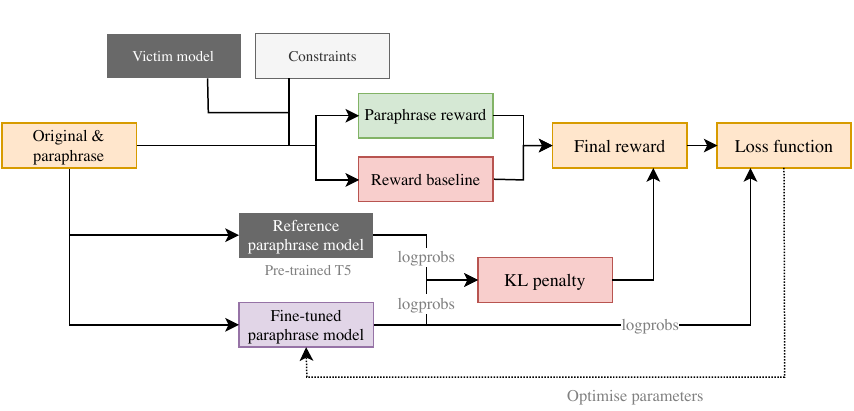}
  \caption{A diagram of the training approach. As input, training uses batches of \textit{(original, paraphrase)} pairs. The parameters are updated using a REINFORCE with baseline algorithm. The overall loss function depends on the reward function, the baseline, the constraints, and the KL divergence penalty, which compares the probabilities computed by the fine-tuned and pre-trained paraphrase models.}
  \label{fig:optimisation}
\end{figure*}

During each training epoch, we generate one paraphrase per original example and collate them into batches of training data. The batches are used to compute a loss function (Section \ref{sec:loss_fn})
, which incorporates both a reward function (Section \ref{sec:paraphrase_reward}) and a baseline (Section \ref{sec:reward_baseline}). We use a set of constraints  (Section \ref{sec:constraints}) to determine if the generated text is valid, and examples that fail receive zero reward. 
Figure \ref{fig:optimisation} shows the overall setup.

During validation, we are able to generate not one, but a whole set of adversarial example candidates per original example by simply using off-the-shelf decoding (Section \ref{sec:decoding}). We call these generated examples the set of \textit{adversarial example candidates}, and consider the attack successful if at least one of them meets the given constraints. Since the limited resource are the available original samples, we compute the attack success rate as the ratio between the number of successful attacks and the number of original examples. Generating more paraphrases can obviously improve the attack success rate, but the generation takes longer and the memory requirements increase. As an effective trade-off between these factors, we have chosen to generate $n=48$ paraphrases per original example. The same procedure is used at test time.
During validation, we also update the reward baseline with the average per-example reward across the candidate set. Training is stopped once the validation set performance improvement drops below a threshold, or after a maximum number of epochs (full details are available in Appendix \ref{sec:hyperparameters}). 



\begin{figure*}[!ht]
\centering
\begin{minipage}{\textwidth}%
\subfloat[][]{
\includegraphics[width=.42\textwidth]{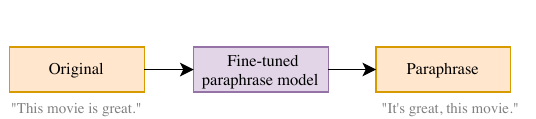}
\label{fig:gen_train}}
\subfloat[][]{
\includegraphics[width=.57\textwidth]{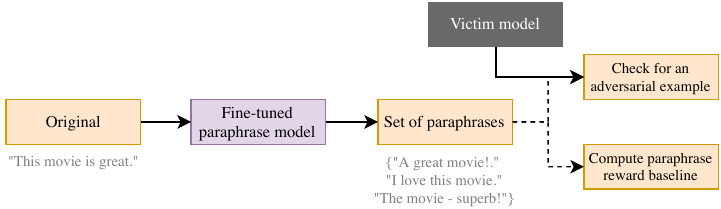}
\label{fig:gen_eval}}
\end{minipage}%
\caption{Sample generation during training and validation. (a) During training, we generate one paraphrase per original example, decoding with nucleus sampling. (b) During validation, we generate a set of paraphrases per original example, decoding with one of four methods (Section \ref{sec:decoding}). We then check if any paraphrase in the set is a successful adversarial example, and also use the set (for the training split) to update the reward baseline (Section \ref{sec:reward_baseline}).}
\label{fig:generation}
\end{figure*}


\subsection{Loss function}
\label{sec:loss_fn}

Let us have an input data distribution $\mathcal{D}$, an original example $x \in \mathcal{D}$, and a pre-trained paraphrase model with parameters $\theta$. Given $x$, the model generates a paraphrase $x'$ with $T$ tokens with probability $p_\theta(x'_{t}|x'_{1},\ldots,x'_{t-1}, x), t = 1 \ldots T$. We note this predictive distribution as $\rho$ for simplicity. 
Our agent attempts to learn a policy $\pi$, still parametric in $\theta$ and initially equal to $\rho$, that can create adversarial examples, for which we have a reward function $r$ that scores success and failure. Training aims to optimise $\pi$ to maximise the expected value of $r$:

$$ \mathbb{E}_\pi (r) = \mathbb{E}_{x \sim \mathcal{D}, x' \sim \pi(x) } r(x,x') $$

To optimise using gradient descent, the gradient $\Delta _\theta \mathbb{E}_\pi (r)$ is required, for which an estimator is provided by the policy gradient theorem \citep{sutton1999policy}:
\vspace{-2pt}
\begin{equation}
\label{eq:grad}
\begin{aligned}
\Delta_\theta \mathbb{E}_\pi (r)= -r \sum\limits_{t=1}^{T} \: \frac{\partial}{\partial \theta} \log \pi(x'_{t}|x'_{1},\ldots,x'_{t-1}, x) \\
\end{aligned}
\end{equation}

\noindent where $x'$ is a sampled sequence (using any of a number of sampling methods). The above is the REINFORCE estimator \citep{williams92reinforce}. Using an  automatic differentiation framework we can convert this into a loss function:
\vspace{-1pt}
\begin{equation}
\label{eq:reinforce}
\begin{aligned}
L_{\scaleto{RF}{3pt}} &= -r \sum\limits_{t=1}^{T} \: \log \pi(x'_{t}|x'_{1},\ldots,x'_{t-1}, x) \\
&= -r \log \pi(x'|x)\\
\end{aligned}
\end{equation} 

\noindent This estimator is unbiased, but it typically exhibits a large variance, which causes slow and unstable learning. The variance can be reduced by subtracting a baseline, $b$, from $r$:
\vspace{-1pt}
\begin{equation}
\label{eq:reinforce_with_baseline}
\begin{aligned}
L_{\scaleto{b}{3pt}} = -(r-b) \log \pi(x'|x)\\
\end{aligned}
\end{equation}
\noindent provided $b$ is highly correlated with $r$. This estimator is biased in the case that $b$ depends on $x'$ \citep{williams92reinforce}, but typically delivers improved training speed and stability. 

Concurrently, we also like to prevent the trained distribution, $\pi$, from diverging too much from the original predictive distribution, $\rho$, since that is likely to affect the coherence and paraphrase quality of the generated text. Following previous work \citep{sequence_tutor, openai_finetuning}, we add a KL divergence penalty, $D_{KL}$, to discourage this behaviour. The modified reward function, after the baseline and the KL divergence term are incorporated, becomes:
\vspace{-2pt}
\begin{equation}
\label{eq:final_reward}
\begin{aligned}
R(x, x') = r(x, x') - b(x) - \beta D_{KL} \\
\end{aligned}
\end{equation}
\noindent where $\beta$ is a scaling constant, and:
\begin{equation}
\label{eq:kl}
\begin{aligned}
D_{KL} = \mathbb{E}_{x \sim \mathcal{D}, x' \sim \pi(x)} [\log \pi(x'|x) - \log \rho(x'|x)]
\end{aligned}
\end{equation}
This leads to the overall loss function:
\vspace{-1pt}
\begin{equation}
\label{eq:loss_fn}
\begin{aligned}
\mathcal{L} = -R(x, x') \log \pi(x'|x)\\
\end{aligned}
\end{equation}

Finally, to prevent longer sequences from being unfairly penalised, we normalise the log probability terms $\log \pi(x'|x)$ and $\log \rho(x'|x)$ in (\ref{eq:kl}) and (\ref{eq:loss_fn}) by dividing each by the generated sequence length, $T$.

\subsection{Paraphrase reward}
\label{sec:paraphrase_reward}

Let $f$ be the probability output by the victim classifier, $x$ be the original example with label $y$, and $x'$ be a paraphrase. Let $V(x, x') = f(x)_y - f(x')_y$ be the degradation in confidence in $y$ that $x'$ induces in $f$. Then the paraphrase reward to use in  (\ref{eq:final_reward}) is:
\begin{equation}
\label{eq:paraphrase_reward}
    r(x, x') =  \max(0, \min(\alpha, \eta \hspace{1pt} \delta(x, x') V(x,x') )) 
\end{equation}

\noindent where $\alpha$ is an upper bound, $\eta$ a scalar multiplier, and $\delta(x,x')$ a Dirac delta function that is 1 if the constraints  (Section \ref{sec:constraints}) are met, and 0 otherwise. 


\subsection{Reward baseline}
\label{sec:reward_baseline}

As shown in Equation \ref{eq:final_reward}, the gradient estimator requires a  baseline $b$ for the reward. We use a per-example baseline, $b(x)$,
and define it as the average reward of the set of adversarial example candidates generated for each $x$ in the training set. The baseline is updated in each validation phase (see Figure \ref{fig:gen_eval}). Intuitively, the $b(x)$ baseline is high when the model can easily generate adversarial examples for $x$, and low when it cannot.

\subsection{Adversarial example constraints}
\label{sec:constraints}
In addition to switching the predicted label, an adversarial example should both preserve meaning \citep{michel-etal-2019-evaluation} and be linguistically acceptable. We enforce these principles by using the following constraints:

\textbf{Label invariance.} The original and paraphrase must have the same ground-truth label. Since the ground-truth label of the paraphrase is latent, this constraint is failed if the paraphrase contradicts the original with a probability $\geq$ 0.2 according to a natural language inference pre-trained model.

\textbf{Is semantically consistent.} The original and paraphrase must have (broadly) the same semantic content. To assess this, we extract sentence embeddings of both using a pre-trained Siamese-BERT model \citep{reimers-2019-sentence-bert}, compute their cosine similarity, and impose a minimum threshold of 0.8. 

\textbf{Is linguistically acceptable.} Paraphrases should be acceptable sentences. This constraint is met only if the generated sentence is deemed linguistically acceptable with a probability $\geq$ 0.5, according to a pre-trained language model. 


Through trial and error, we decided to also introduce two  additional constraints to prevent the generation of undesirable solutions: 

\textbf{The sentence length is similar.} To prevent the generation of very short sentences, we require the original and paraphrase to have sentence length within 30 characters of each other.

\textbf{Avoids linking contrast phrases.} Regardless of the true class, the model can ``soften'' the generated paraphrase by starting or ending it with a linking contrast phrase, such as ``however'' or ``nonetheless'' (see Table \ref{tab:transformations}). To encourage the generation of more interesting solutions, we disallow this behaviour, unless the original example itself starts or ends with that phrase.

\section{Experimental setup}
\label{sec:experimental_setup}

\subsection{Datasets} 
\label{sec:datasets}

The experiments have been carried out on two English sentiment analysis datasets, each consisting of sentences or short text fragments. The first is the Rotten Tomatoes dataset \citep{rotten_tomatoes} which contains extracts of movie reviews with sentiment labelled as either positive or negative. We have used the predefined training, test and validation splits. The second is the Financial PhraseBank dataset \citep{financial_phrasebank} which contains financial news fragments with sentiment labelled as positive, neutral or negative. We have used the dataset version with at least 50\% annotator label agreement, and randomly selected 10\% of the data as the validation set and 10\% as the test set. 

For both datasets, we have excluded the training examples that the victim model classified incorrectly, as they could be said to be already ``adversarial''. We have also only included examples with 32 tokens or fewer, since the pre-trained paraphrase model had been trained on sequences in that range. The final datasets are relatively small and, as such, are challenging to train on effectively. 

\subsection{Hyperparameters and design choices}
\label{sec:decoding}

Since design choices significantly impact the attack success rate of the fine-tuned model, in the experiments we have explored the impact of two: the decoding sampling temperature used during training, which controls the exploration of the agent; and the decoding method used for inference and evaluation, which affects the diversity and quality of the generated candidate set. All other hyperparameters have been kept constant. Appendix \ref{sec:hyperparameters} provides a complete list of the hyperparameters and more training details. 

\subsubsection{Decoding temperature during training} During training, we generate the paraphrases using nucleus sampling, with the probabilities returned by a softmax operator. The temperature parameter used in the softmax, which we denote as $\tau$, visibly affects the generated text: higher temperatures produce more randomness and more diverse training examples, but also a lower percentage of valid English sentences. We have therefore investigated two values of $\tau$: a low-temperature condition of $\tau$ = 0.85, and a high-temperature condition of $\tau$ = 1.15. We found both these values retain sentence semantics and give somewhat diverse examples, while also being fittingly different from each other. 

\subsubsection{Decoding method during evaluation}

During evaluation, i.e. both validation and test, we generate a set of $n=48$ paraphrases per original example, and then check if any are valid adversarial examples based on the constraints. The decoding method used influences the characteristics of the generated text --- and consequently, the attack success rate. This is a key design choice and we have therefore investigated four different methods: 

\begin{enumerate}
\item \textit{Sampling.} We have used nucleus sampling, with top-p at 0.95 and the temperature at 1.

\item \textit{Beam search.} We have set the number of beams to 48, one per generated example.

\item \textit{Low-diversity beam search.} Diverse beam search \citep{diverse_beam_search} is a beam-search variant that increases diversity of generated sequences by dividing the beams into groups and encouraging diversity between them. For this condition we have used six beam groups, set the diversity penalty to 1, and again used 48 beams, one per generated example.  

\item \textit{High-diversity beam search.} As above, but with 48 beam groups instead of six.
\end{enumerate}

\section{Results}
\label{sec:results}

\subsection{Attack success rate}
\label{sec:trained_vs_untrained}
First, we have investigated if the proposed approach improved the attack success rate over the original paraphrase model. As mentioned in Section \ref{sec:overview}, an attack is counted as successful if at least one example in the candidate set is validated as a true adversarial example, in that it meets the validity constraints while also inducing a misclassification (NB: typically, many more than one are). The attack success rate is then computed as the percentage of successful attacks.
In this section, we report the attack success rate for the original and fine-tuned paraphrase models on the test split of the two datasets, across the two temperature settings and the four decoding methods. For all conditions, we have run three runs from different random seeds and chosen the best performing on the validation set.


\textbf{Results.} The results are reported in Table \ref{tab:main_results}, showing that the proposed approach has improved the attack success rate across all training conditions. All improvements have been statistically significant ($p<0.01$) according to a bootstrap test, as recommended by \citet{dror2018}. We have found no clear best between the two temperature values ($\tau=0.85$ and $\tau=1.15$), but beam search as the decoding method has reported the highest success rates on both datasets (61.9\% on Rotten Tomatoes and 82.0\% on Financial PhraseBank). We have also observed a substantial variance in the results across random seeds (see Figure \ref{fig:eval_perf}). This problem is well known in reinforcement learning \citep{henderson2017deeprl} and may be mitigated by more sophisticated baselines, such as RELAX \citep{grathwohl2018RELAX}.
We have also performed a run on the Rotten Tomatoes dataset by including examples with up to 48 tokens instead of 32, and found no noticeable difference in attack performance.


\begin{table*}[!ht]
\centering
\small
\resizebox{0.9\textwidth}{!}{%
\begin{tabular}{@{}lccccccc@{}}
\toprule
                                         & \multicolumn{3}{c}{\textit{Rotten Tomatoes}} &  & \multicolumn{3}{c}{\textit{Financial PhraseBank}} \\ \midrule
\textbf{Decoding method} &
  \textit{Original} &
  \textit{$\tau=0.85$} &
  \textit{$\tau=1.15$} &
  \textit{} &
  \textit{Original} &
  \textit{$\tau=0.85$} &
  \textit{$\tau=1.15$} \\
  
%
%


%
%
\textit{Sampling} & 30.6         & 41.8         & 46.8       &  & 20.8          & 72.3          & 83.6          \\
\textit{Beam search}                             & 14.5           & \textbf{85.5}          & \textbf{85.5}        &  & 11.3           & 82.4          & \textbf{88.1}        \\
\textit{Low-diversity beam search}   & 20.6         & 43.2         & 57.7        & & 13.2         & 79.3         & 79.9          \\
\textit{High-diversity beam search}   & 24.5         & 66.0         & 37.3        &  & 21.4         & 76.1         & 77.4         \\ \bottomrule

\end{tabular}%
}
\caption{Test-set attack success rate (as a percentage) of the original and fine-tuned models. We use $\tau$ to refer to the decoding temperature during fine-tuning. 
The best results for each dataset are in bold. Fine-tuning has improved the attack success rate for all conditions ($p < 0.01 $ according to a bootstrap hypothesis test \citep{dror2018}). Amongst the evaluation decoding methods, beam search has had the highest success rates, followed by low-diversity beam search.}
\label{tab:main_results}
\end{table*}

\subsection{Comparison with established adversarial attacks} 
\label{sec:trained_vs_tokenmod}

Next, we have compared the proposed approach to results from a range of competitive adversarial attacks. We have created six different attacks using the TextAttack library \citep{textattack}, varying the transformations and the search method, and we also compared against three established adversarial attack recipes: TextFooler \citep{Jin2020TextFooler}, BAE-R \citep{BAE} and IGA \citep{IGA}. These attacks use a variety of language models and embeddings, span a range of query budgets (i.e., the number of victim model queries per example attacked), and have a corresponding range of attack success rates \citep{yoo2020searching}. To ensure a fair comparison, each attack abides by the same constraints as the fine-tuned model, while during the iterations the search is not allowed to modify the same word twice or stopwords 
(more details are provided in Appendix \ref{sec:token_modification_algorithms}). For performance comparison, we have used the best-performing fine-tuned model from Table \ref{tab:main_results} for each dataset, and we have compared it with these attacks in terms of both attack success rate and number of victim model queries.


\textbf{Results.} The results in Table \ref{tab:token_modification_comparison} show that the fine-tuned model has achieved a much higher attack success rate than the corresponding adversarial attacks for a comparable average number of queries (e.g., 85.5\% with 48 queries vs 67.7\% with 52 for Rotten Tomatoes). In fact, its attack success rate has been similar to that of the most query-expensive attack tested, despite requiring a fraction of its queries (48 vs 976 on average for Rotten Tomatoes and 1331 for Financial PhraseBank). More so, the fine-tuned model has been able to generate not one, but many successful adversarial examples per original, averaging 19.6 out of 48 for Rotten Tomatoes and 27.6 for Financial PhraseBank. This staggering difference directly stems from the inherent design advantages of the seq2seq approach.


\begin{table*}[!ht]
\centering
\small
\resizebox{0.9\textwidth}{!}{%
\begin{tabular}{@{}lccccc@{}}
\toprule
                       & \multicolumn{2}{c}{\textit{Rotten Tomatoes}} &  & \multicolumn{2}{c}{\textit{Financial PhraseBank}} \\ \midrule
\textbf{Attack method} & Success \%          & Avg Queries         &  & Success \%            & Avg Queries            \\ \midrule
\textit{Adversarial attacks (see Appendix \ref{sec:token_modification_algorithms})} &      &     &  &      &     \\ \midrule
LM-based, low-budget   & 39.6  & 42 &  & 39.0  & 58 \\
Embedding-based, low-budget (TextFooler) & 67.7 & 52 & & 50.9 & 72 \\
LM-based, low-budget (BAE-R)  & 65.2 & 53 & & 60.4 & 76 \\ 
LM-based, medium-budget (variant 1)    & 69.4  & 282 &  & 69.2  & 290 \\
Embedding-based, medium-budget                    & 79.4  & 327 &  & 64.8  & 498 \\
LM-based, medium-budget (variant 2)         & 69.1  & 425 &  & 69.2  & 463 \\ 
 LM-based, high-budget  (variant 1)   & 77.2  & 790 &  & 75.5  & 697 \\
Embedding-based, high-budget (IGA)  & 86.9 & 512 & & 76.1 & 860 \\ 
LM-based, high-budget (variant 2)    & 93.9  & 976 &  & 91.2  & 1331 \\
\midrule
\textit{Fine-tuned model}              &      &  Queries (Avg Successes) &  &      &  Queries (Avg Successes)   \\ \midrule
Eval: beam search, train: $\tau=1.15$, best run                                 & 85.5 & 48 (19.6) &  & 88.1  & 48 (27.6) \\
\bottomrule
\end{tabular}%
}
\caption{Comparison of the best-performing fine-tuned models against a range of other adversarial attacks. LM: language model; variants 1, 2: see Appendix \ref{sec:token_modification_algorithms}. For these attacks we show the average number of victim model queries needed to find a single adversarial example. In contrast, the proposed fine-tuned model performs a fixed number of victim model's queries (48) and generates multiple adversarial examples. The rows are sorted by increasing Avg Queries value for the Rotten Tomatoes dataset. 
}
\label{tab:token_modification_comparison}
\end{table*}




\section{Human validation of the adversarial examples}
\label{sec:human_validation}
To assess if the proposed approach maintains label invariance, we have performed a small-scale human validation. 

The assessment has used samples from the Financial PhraseBank test set and compared adversarial examples from the fine-tuned model, the original paraphrase model, and the most successful compared adversarial attack. Details of the methodolgy of the evaluation are given in Appendix \ref{sec:human_methodology}.

\textbf{Results.} 
The fine-tuned model and the compared attack have been able to retain the same ground-truth label as the original example at approximately similar rates (59\% and 50\%, respectively). The original paraphrase model has been able to retain the ground-truth label in all cases (as expected) but at the price of a drastically lower success rate.



 The results show that the fine-tuned model has degraded paraphrasing capability compared to the original paraphrase model, but it still has achieved higher label invariance than the compared approach. These results should be judged alongside the success rate, which has been approximately 8 times higher for the fine-tuned model than the original paraphrase model on the Financial PhraseBank test set (Table \ref{tab:main_results}). It is also worth noting that label invariance rates vary considerably in the literature, foremost because there are no ``standard settings'' to measure them across papers. The rates depend on many factors, such as the instructions given to the evaluators,  their harshness or lenience, the use of crowd-sourcing platforms, the datasets used, etc. The most important requirement is that the evaluation is applied equally to all compared approaches, as we do here.

\section{Discussion}
\label{sec:analysis}

\subsection{Extending to other text classification tasks}

To see if our approach generalises to classification datasets outside sentiment analysis, we have tested the approach on the Text REtrieval Conference (TREC) question-type classification dataset \citep{TREC}. This dataset classifies questions as one of six categories, each labelled as the type of answer that is expected (e.g. \textit{Person}, \textit{Description}). For this experiment, we have used the best set of experimental parameters identified in Section \ref{sec:results} and Table \ref{tab:main_results}, i.e., beam search decoding and temperature of 1.15. We have also tested the approach against two different victim classifiers, the first using a BERT model \citep{BERT} and the second a DistilBERT one \citep{DistilBERT}, each fine-tuned on the given dataset before being subjected to the attacks. Each experiment has been run from three different random seeds and the average results are reported. For performance comparison, we have compared our approach against two probing token-modification attacks from Table \ref{tab:token_modification_comparison}: the first (LM-based, low-budget (BAE-R)) is an attack that requires a comparable amount of victim model queries, and the second (LM-based, high-budget (variant 2)) is the highest-performing of the compared approaches from Table \ref{tab:token_modification_comparison}. As a baseline/lower bound, we have also included the original paraphrase model (Untrained model). The results are presented in Table \ref{tab:ablation_results}, showing that our model has been able to achieve the highest success rates and the largest amounts of successful adversarial examples per query.  In addition, the trends have not been significantly affected by the choice of victim model. We have also performed a run by including input examples with up to 48 tokens instead of 32, and found no noticeable difference in performance. Overall, this gives evidence to the ability of our approach to generalise well to classification tasks other than sentiment analysis, and across different base models.

\begin{table*}[!ht]
\centering
\small
\resizebox{0.9\textwidth}{!}{%
\begin{tabular}{@{}lccccc@{}}
\toprule
                       & \multicolumn{2}{c}{\textit{BERT}} &  & \multicolumn{2}{c}{\textit{DistilBERT}} \\ \midrule
\textbf{Attack method} & Success \%          & Avg Queries         &  & Success \%            & Avg Queries            \\ \midrule
\textit{Adversarial attacks (see Appendix \ref{sec:token_modification_algorithms})} &      &     &  &      &     \\ \midrule
LM-based, low-budget (BAE-R)  & 88.5 & 31.8 & & 89.3 & 30 \\ 
LM-based, high-budget (variant 2)    & 91.2  & 150.1 &  & 94.1  & 129.6 \\
\midrule
\textit{Generative models}              &      &  Queries (Avg Successes) &  &      &  Queries (Avg Successes)   \\ \midrule
Untrained model & 42.2 & 48 (1.9) &  & 47.38  & 48 (2.3) \\
Fine-tuned model & 95.8 & 48 (26.6) &  & 97.9  & 48 (28.3) \\
\bottomrule
\end{tabular}%
}
\caption{Comparison of the proposed model against two probing approaches and the untrained paraphrase model over the TREC dataset. The victim classifiers use BERT and DistilBERT, respectively (top-level column headings), and the results are the average of three runs from three different random seeds. The results show that our model has achieved the highest success rates and the largest amounts of successful adversarial examples per query.  
}
\label{tab:ablation_results}
\end{table*}

In addition, in Table \ref{tab:ablation_examples} we present three qualitative examples of attacks generated by the various approaches. The examples show that the two compared token-modification approaches have somehow distorted the original meaning and compromised linguistic acceptability in some cases, while the proposed model has been able to retain the meaning more closely and generate linguistically-correct attacks in all cases. In addition, in the second example the proposed model has been to generate a much more articulate transformation of the original sentence than those in reach of the two token-modification approaches.

\input{table_ablation_examples}

\subsection{Decoding method analysis} 
\label{sec:analysis_decoding_method}

In this section we have compared the different decoding methods used during evaluation/inference in terms of attack success rate, diversity and fluency. Examples of the generated text are in Appendix \ref{sec:examples}.

\textbf{Preprocessing.} 
To measure diversity and fluency, we have selected the two runs with the highest attack success rate per decoding condition across both training temperatures.\footnote{We select the best two because there is significant variation in runs between random seeds and it would not be beneficial to compare models that have failed to train properly.}

\begin{figure*}[!ht]
\centering
\begin{minipage}{0.95\textwidth}%
\subfloat[][]{
\includegraphics[width=0.45\linewidth]{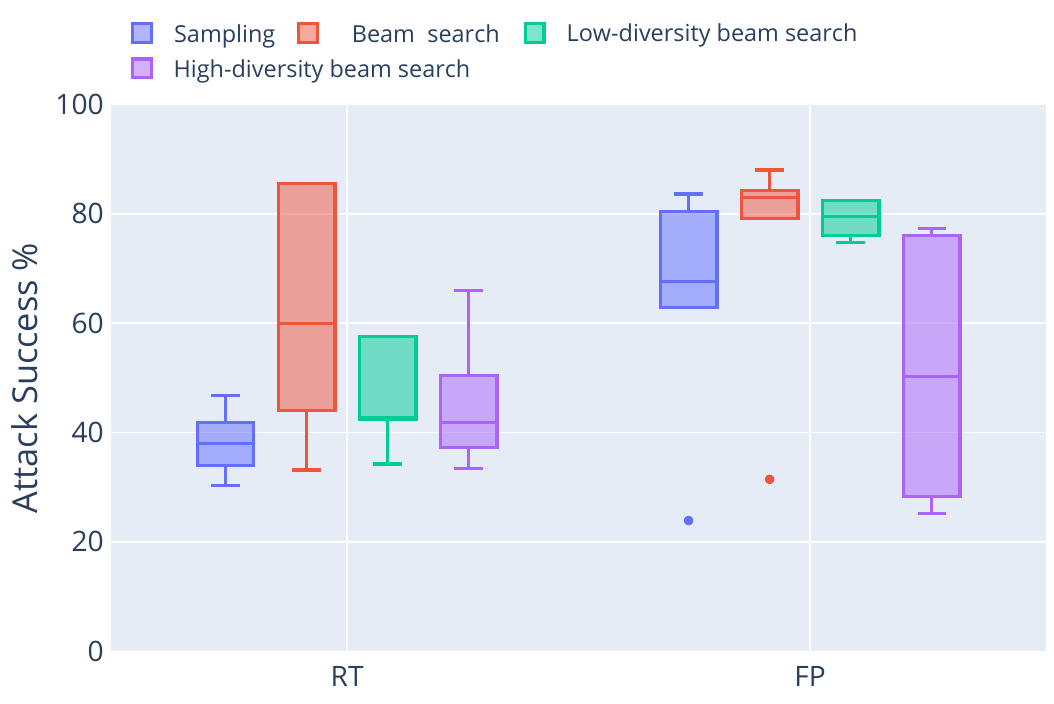}
\label{fig:eval_perf}}
\qquad
\subfloat[][]{
\includegraphics[width=0.45\linewidth]{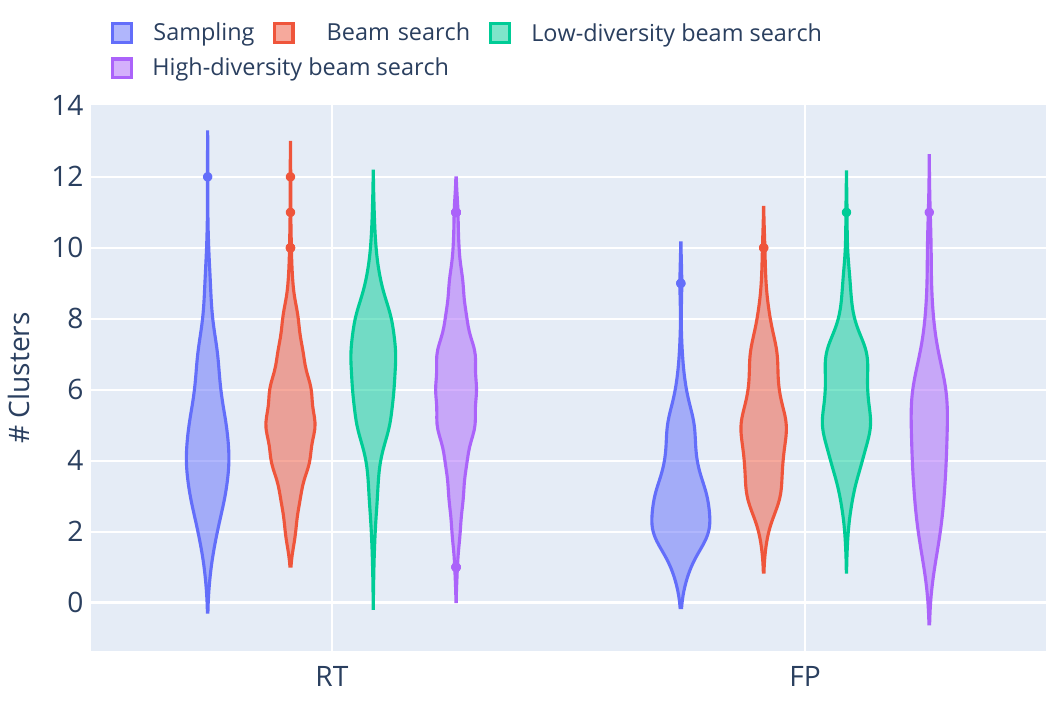}
\label{fig:eval_n_clusters}
}
\end{minipage}%
\caption{Attack success rate and diversity of decoding methods. For each graph: RT = Rotten Tomatoes, FP = Financial PhraseBank. (a) Attack success rate by decoding evaluation method, across seeds.  We see the common RL training problem of high variance across seeds \citep{henderson2017deeprl}. Beam search and low-diversity beam search perform best, on average. (b) Candidate set diversity of each decoding method, which we measure using a cluster-based score (see Section \ref{sec:analysis_decoding_method}). 
More clusters indicates a more diverse candidate set. }
\label{fig:decoding_attack_and_diversity}
\end{figure*}

\textbf{Attack success rate.} We found that beam search had the highest attack success rates, followed by low-diversity beam search, although we observed significant variation between runs for all methods (Figure \ref{fig:eval_perf}).

\textbf{Diversity.} To quantify the diversity, we have built a score using the same procedure used for the selection of the adversarial examples for the human validation, described in full in Section \ref{sec:filter_human}. The score has been simply defined as the number of distinct clusters returned by HDBSCAN, with the individual examples not included in any cluster counted as clusters themselves. The results are shown in Figure \ref{fig:decoding_attack_and_diversity}. 

\begin{figure*}[!ht]
\centering
\begin{minipage}{\textwidth}%
\subfloat[][]{
\includegraphics[width=0.45\linewidth]{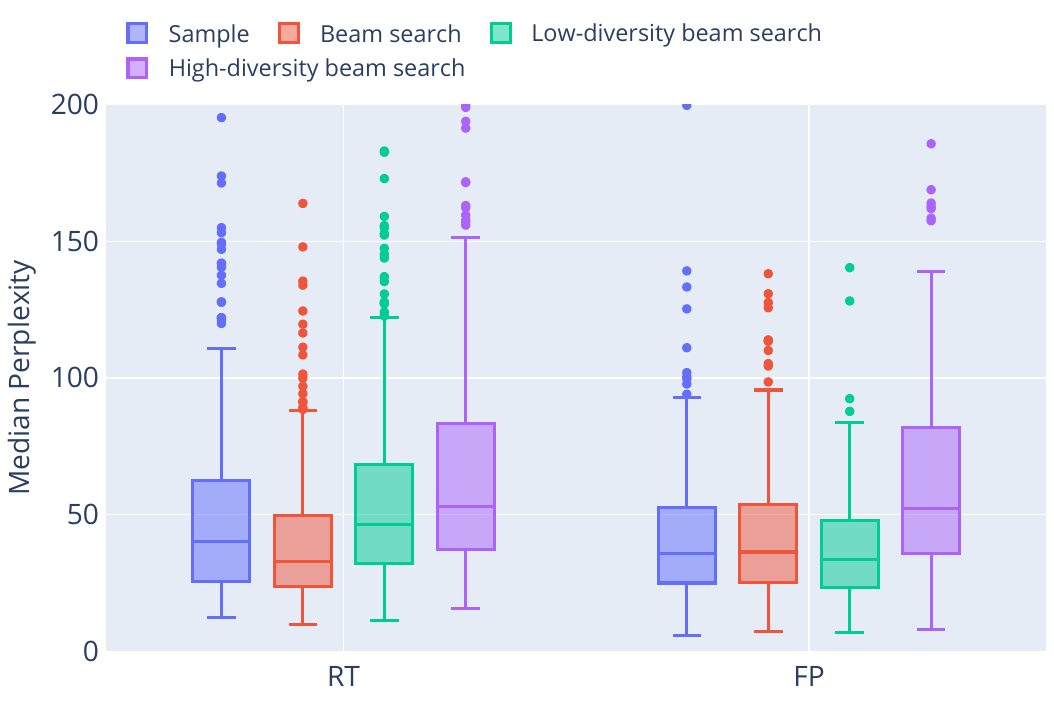}
\label{fig:eval_perplexity}}
\qquad
\subfloat[][]{
\includegraphics[width=0.45\linewidth]{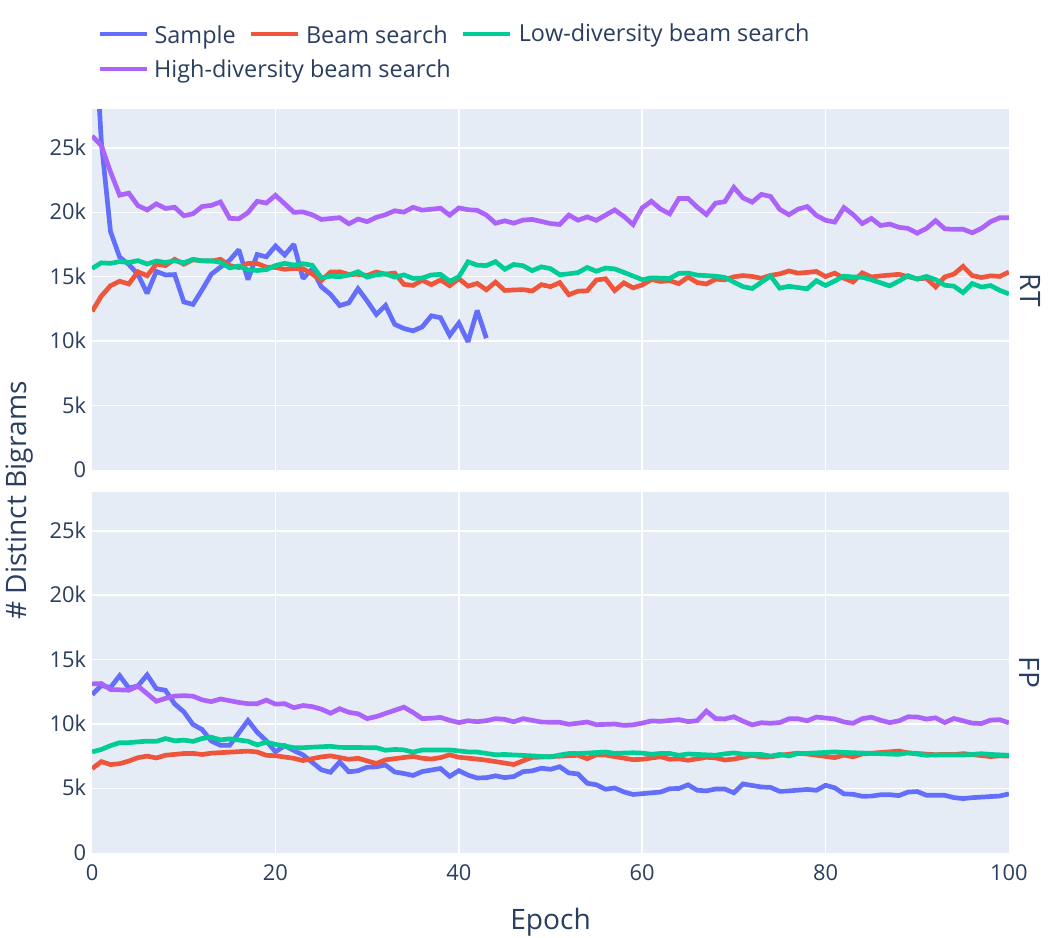}\label{fig:eval_bigrams}}
\end{minipage}%
\caption{Fluency scores for the various decoding methods. RT = Rotten Tomatoes, FP = Financial PhraseBank. (a) Median perplexity of the generated candidate sets, with examples combined from the top two runs of each decoding method. Three of the methods have been approximately comparable, while high-diversity beam search has consistently produced the least fluent candidates. (b) Average number of distinct bigrams generated per epoch performing evaluation on the training set. High-diversity beam search (in purple) has consistently generated more unique bigrams than the other methods. The \textit{sampling} decoding method has displayed a marked decrease in diversity along the epochs, while the others have remained approximately constant. These results confirm the expected trade-off between fluency and diversity.}
\label{fig:decoding_fluency}
\end{figure*}

Overall, low-diversity beam search has generated the most diverse examples, while sampling has generated the least (Figure \ref{fig:eval_n_clusters}).  We have found that, somewhat surprisingly, the high-diversity beam search showed less diversity than the low-diversity beam search, probably because the method tended to generate many ``degenerate'' examples that were clustered together in our clustering procedure. We have also found that the sampling method generated fewer unique examples on average than the other conditions. We speculate that the token probability over the vocabulary tends to become more concentrated as fine-tuning progresses, as supported by Figure \ref{fig:eval_bigrams}. 

\textbf{Fluency.} As commonly done (e.g. \citet{wang2018sentigan}), we have used language model perplexity as a proxy for fluency. We have also used the number of unique generated bigrams 
performing evaluation on the training set after every epoch as a further measurement of diversity. The results are summarised in Figure \ref{fig:decoding_fluency} and show that overall, fluency was similar for all methods except high-diversity beam search.  


\textbf{Overall.} Finding the most effective method for this task is subjective, but it can be argued that low-diversity beam search has achieved the best results by balancing attack success rate, diversity, and fluency.




\subsection{Learned strategies}
\label{sec:reward_design}

During training, the model has displayed a wide range of different generating behaviours, such as phrase shuffling, synonym swapping and double negatives, a selection of which is shown in Table \ref{tab:transformations}. Some behaviours have proved compatible with our validity constraints, others have violated them.
In the initial stages of our research, we had to adjust the reward function repeatedly to disallow some unwanted behaviours, leading to the additional constraints of Section \ref{sec:constraints}.\footnote{This analysis was carried out solely on the training set.} 
Some behaviours, such as ignoring grammar, can be considered exploitations of the individual components of the reward function (i.e., ``reward hacking''). In addition, the constraints have not been able to perfectly filter all the actual adversarial examples.
However, since the components and the constraints are simply proxies for human preferences, the reward can still be effective so long that these behaviours remain limited. 







\begin{table*}[!ht]
\centering
\small
\resizebox{\textwidth}{!}{%
\begin{tabular}{@{}ll@{}}
\toprule
\textbf{Transformation} &
  \textbf{Example} \\ \midrule
  Original &
  safe conduct, however ambitious and well-intentioned, fails to hit the entertainment bull’s-eye \\
Genuine paraphrase &
  safe conduct might be ambitious and well-intentioned, but it misses the entertainment bull’s-eye \\
Synonym swapping &
  safe conduct, however ambitious and well-intentioned, fails to strike the   entertainment bull’s-eye \\
Case changes &
  safe Conduct, however Ambitious And well-intentioned, fails to hit the   entertainment bull’s-eye \\
Adding/removing punctuation &
  safe conduct however ambitious and well-intentioned, fails to hit the   entertainment bullseye..:;: \\
Ignoring grammar &
  safe conduct ambitious well-intentioned bull’s-eye \\
Phrase shuffling &
  safe conduct, fails to hit the entertainment bull’s-eye, however   ambitious and well-intentioned \\
Contradictions &
  safe conduct, however ambitious and well-intentioned, hit the   entertainment bull’s-eye \\
Sentence truncation &
  safe conduct, however ambitious and well-intentioned \\
Padding to max length &
  safe conduct, however ambitious and well-intentioned, fails to hit the   entertainment bull’s-eye and and and and and and \\
Very short sentences &
  safe conduct. \\
Inserting phone-numbers &
  safe conduct, however ambitious and well-intentioned, fails to hit the   entertainment bull’s-eye 888-739-5110 888-739-5110 \\
Using Unicode characters &
    safe cond©$\sqrt{}$t, however ambitious and well-intentioned, fails to hit the entertainment bull’s-eye. \\
Adding linking contrast phrases &
  although safe conduct, however ambitious and well-intentioned, fails to   hit the entertainment bull’s-eye, nonetheless \\
Repeating phrases &
  safe conduct, however ambitious and well-intentioned, fails to hit the   entertainment bull’s-eye, entertainment bull’s-eye\\
Change language of fragment & 
    safe conduct, however ambitieux et well-intentioned, fails to hit the entertainment bull’s-eye \\
Rhetorical question &
  safe conduct, however ambitious and well-intentioned, fails to hit the   entertainment bull’s-eye - but why? \\
Double negatives &
  safe conduct, however ambitious and well-intentioned, fails to fail to   hit the entertainment bull’s-eye \\ \bottomrule
\end{tabular}%
}
\caption{Examples of the model's generating behaviours during training. Rather than only paraphrasing, the model has exhibited a variety of different behaviours. We have introduced two additional constraints to disallow the behaviours of generating very short sentences and adding linking contrast phrases, but we have allowed the others. 
}
\label{tab:transformations}
\end{table*}

\section{Conclusion}
\label{sec:conclusion}

This paper has proposed an approach for generating adversarial attacks for a text classifier based on fine-tuning a seq2seq paraphrase model. The proposed approach fine-tunes the paraphrase model with a novel reward function that encourages misclassifications in the victim model while simultaneously ensuring that the generated attacks abide by a set of validity constraints. The experimental results over three datasets and two victim classifiers have shown that the fine-tuned model has been able to produce many more adversarial examples than the original paraphrase model. It has also proved much more effective and efficient than a set of compared attacks in terms of success rate and number of successful attacks/number of queries trade-off.
Future work could include exploring more efficient and stable training algorithms, incorporating actual human preferences into the reward objective,
and experimenting with different pre-trained seq2seq models, such as those for style transfer or dialogue generation.

\bibliography{custom}
\bibliographystyle{ACM-Reference-Format}

\clearpage
\appendix

\section{Training details}
\label{sec:hyperparameters}
This section provides all the details of our training setup. 

We have used an AdamW optimiser with early stopping, using the attack success rate of the validation set as the metric, and stopping once the metric dropped below the running median, or when a maximum number of epochs was reached (100 for Rotten Tomatoes, 200 for Financial PhraseBank). We have not done any layer freezing during training as we noticed that it tended to reduce performance. 

After the sample selection from the datasets, the splits contained the following number of examples: Rotten Tomatoes --- training: 2972, validation: 367, test:  359; Financial PhraseBank --- training: 1370, validation: 167, test: 159. The hyperparameters that have been kept constant across all experiments are listed in Table \ref{tab:hyperparameters}. The details of the various models --- paraphrase model, victim models, and reward component models --- are given in Table \ref{tab:models}. 

\begin{table}[!ht]
\centering
\small
\begin{tabular}{@{}ll@{}}
\toprule
\textbf{Hyperparameter}                         & \textbf{Value}     \\ \midrule
\textit{General}                                &                    \\ \midrule
LR                                              & $1 \times 10^{-4}$ \\
Batch size                              & 32                 \\
Gradient accumulation steps                     & 2                  \\
Max paraphrase length                       & 48                 \\
Min paraphrase length                       & 3                  \\
Max original length                         & 32                 \\
Padding multiple                                & 8                  \\ \midrule
\textit{Training generation}                    & \textbf{}          \\ \midrule
Generated sequences per original                           & 1 \\
Top-p                                   & 0.95               \\ \midrule
\textit{Eval generation}                        &                    \\ \midrule
Generated sequences per original                            & 48                 \\
Top-p (S)                                & 0.95               \\
Temperature (S)                          & 1                  \\
Number of beams (BS, DBS) & 48                 \\
Diversity penalty (DBS)         & 1                  \\ \midrule
\textit{Reward function}                        &                    \\ \midrule
Reward bounds                                   & [0, $\alpha$ = 10]             \\
Victim degradation multiplier ($\eta$)                           & 35                 \\

$D_{KL}$ scaling coefficient ($\beta$)            & 0.4 \\
Character difference threshold                       & $\pm 30$                 \\
Cosine similarity threshold                                   & 0.8 ($\geq$)       \\
NLI contradiction threshold                         & 0.2 ($\leq$)       \\
Linguistic acceptability threshold                         & 0.5 ($\geq$)       \\ \bottomrule
\end{tabular}%

\caption{Hyperparameters used for training and evaluation across all experiments. S refers to sampling, BS to beam search, DBS to the two diverse beam search conditions, and NLI to natural language inference.
The signs for the thresholds indicate the direction needed to meet the condition. 
}
\label{tab:hyperparameters}
\end{table}

\begin{table*}[!ht]
\centering
\small
\begin{tabular}{@{}lll@{}}
\toprule
\textbf{Purpose} & \textbf{Size} & \textbf{Identifier}                                 \\ 
\midrule
Paraphraser                 & 892 & \href{https://huggingface.co/prithivida/parrot\_paraphraser\_on\_T5}{prithivida/parrot\_paraphraser\_on\_T5}  \citep{prithivida2021parrot}        \\
Victim (RT) & 268 & \href{https://huggingface.co/textattack/distilbert-base-uncased-rotten-tomatoes}{textattack/distilbert-base-uncased-rotten-tomatoes} \\ 
Victim (FP) & 329 & \href{https://huggingface.co/mrm8488/distilroberta-fine-tuned-financial-news-sentiment-analysis}{mrm8488/distilroberta-fine-tuned-financial-news-sentiment-analysis}  \\
Victim (TREC) - BERT & 433 & \href{https://huggingface.co/aychang/bert-base-cased-trec-coarse}{aychang/bert-base-cased-trec-coarse}  \\
Victim (TREC) - DistilBERT & 263 & \href{https://huggingface.co/aychang/distilbert-base-cased-trec-coarse}{aychang/distilbert-base-cased-trec-coarse}  \\
Acceptability  & 47   & \href{https://huggingface.co/textattack/albert-base-v2-CoLA}{textattack/albert-base-v2-CoLA}                 \\
STS &  134 & \href{https://huggingface.co/sentence-transformers/paraphrase-MiniLM-L12-v2}{sentence-transformers/paraphrase-MiniLM-L12-v2} \\
Contradiction       & 54         & \href{https://huggingface.co/howey/electra-small-mnli}{howey/electra-small-mnli} \\   
\bottomrule
\end{tabular}%
\caption{The models used in this paper. We used small, distilled models to increase training speed and because of our GPU memory requirements, but larger models would give better performance. As above, RT stands for Rotten Tomatoes, FP for Financial PhraseBank. All identifiers refer to models on the \href{https://huggingface.co/models}{Hugging Face Model Hub}. Size is given in MB. }
\label{tab:models}
\end{table*}

The GPU model used for training has been an NVIDIA Quadro RTX 6000 with 24GB RAM. Each training run has used a single GPU and the runtime varied from around 4 hours to around 64. The runtime depended heavily on the decoding method (\textit{sampling} was the fastest) and the maximum number of epochs. The training could be potentially sped up by reducing the number of the generated paraphrases during validation. In addition, using a GPU with more memory would allow increasing the batch size, which would also reduce the  training time considerably. Occasionally, for some random seeds and conditions, we have run out of memory on the GPU, most likely due to unusually long predictions; repeating the runs with different seeds has always circumvented this issue.



\section{Compared attacks}
\label{sec:token_modification_algorithms}

We have created a variety of attack types using the TextAttack library of \cite{textattack}. For all attacks, we have set the goal function as untargeted classification (i.e., attacking the correct label with any other class).  We have varied the transformation, the search method, and some hyperparameters of the algorithms, and used the sets of constraints of Section \ref{sec:constraints}. We have also used two extra constraints commonly used to improve the effectiveness of the search (e.g. \citep{seq2sick, Jin2020TextFooler}): the attacks have not been allowed to modify the same word twice, nor modify stopwords. Tables \ref{tab:models} and \ref{tab:attack_codes} provide full details.

\begin{table*}[!htb]
\centering
\resizebox{\textwidth}{!}{%
\begin{tabular}{@{}lllll@{}}
\toprule
\textbf{Attack identifier} & \textbf{Transformations}                                  & \textbf{Search method} & \textbf{Max candidates} & \textbf{Parameters} \\ \midrule

LM-based, low-budget    & Word replacement using language model & Beam search & 5  & $bw=2$  \\
Embedding-based, low-budget (TextFooler) & Word replacement using CF embedding & WIR (delete) & 50 & - \\
LM-based, low-budget (BAE-R) & Word replacement using language model & WIR (delete) & 50 & -  \\
LM-based, medium-budget (variant 1) & Word replacement using language model & Beam search & 25 & $bw=5$  \\
Embedding-based, medium-budget  & Word replacement using CF embedding   & Beam search & 25 & $bw=5$  \\
LM-based, medium-budget (variant 2)  & Word replacement using language model                     & Genetic algorithm      & 25                      & $p=60$, $mi=20$, $mr=5$   \\
 LM-based, high-budget  (variant 1) & Word replacement using language model & Beam search & 50 & $bw=10$ \\
Embedding-based, high-budget (IGA) & Word replacement using CF embedding & Genetic algorithm & 50 & $p=60$, $mi=20$, $mr=5$ \\
LM-based, high-budget (variant 2)     & Word replacement, addition, deletion using language model & Beam search            & 25                      & $bw=5$       \\
 \bottomrule
\end{tabular}%
}
\caption {Compared attacks and their parameters. All attacks were generated using the TextAttack package. We have used a DistilRoBERTa language model\textsuperscript{*}. LM refers to the language model, and CF embedding refers to GloVE word embeddings that have been counter-fitted, a procedure introduced in \citet{mrksic2016CounterFitting} and commonly used in finding word replacements. WIR (delete) stands for Word Importance Ranking by deletion, a search method where the importance of each token is estimated by deleting it and measuring the true-class confidence drop in the victim model. We use $bw$ to mean beam width, and for the genetic algorithm search parameters, $p$ is the population size, $mi$ the maximum number of iterations, and $mr$ the maximum number of replacements per index.}
\small\textsuperscript{*} \url{https://huggingface.co/distilroberta-base}

\label{tab:attack_codes}
\end{table*}

\section{Human validation methodology}
\label{sec:human_methodology}

In this section we present the methodology used for the human validation. 

\textbf{Annotators.} The validation has used three annotators, one of which was a native English speaker and two were proficient English second-language speakers.\footnote{And, for full disclosure, co-authors of this submission. For this reason, the validation has been carried out \textit{blindly}, without knowledge of the generating approach of the examples under assessment.}

\textbf{Models.} We have compared the original paraphrase model, the fine-tuned model, and the best-performing baseline algorithm (\textit{LM-based, high-budget (variant 2)} from Table \ref{tab:token_modification_comparison}) based on the attack success rate. For the fine-tuned model we have chosen a run using low-diversity beam search, since in Section \ref{sec:analysis_decoding_method} we argue that this has proved the best decoding method overall.  

\textbf{Dataset.} We have used 41 original examples from the Financial PhraseBank test set and generated adversarial examples for them. The original examples were split into two groups. The first is where all the three methods were able to produce adversarial examples (18 examples). In the second, the fine-tuned model and the baseline algorithm were able to produce adversarial examples, but the original paraphrase model was not (23 examples). The original examples have been selected so as to have a close-to-balanced class distribution. 

\begin{figure*}[!ht]
	\centering
	\includegraphics[width=0.9\linewidth]{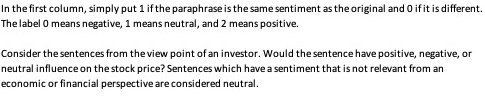}
	\caption{Screenshot of the instructions provided to the annotators for the human validation.}
	\label{fig:human_instructions}
\end{figure*}

\textbf{Annotator instructions.} The aim of the validation has been to assess whether the adversarial examples generated by the three approaches had retained the same ground-truth label as the original, and this question was asked to the annotators (Figure \ref{fig:human_instructions} shows a screenshot of the instructions). To determine the sentiment, we have used the same instructions as the original annotation of the Financial PhraseBank dataset \citep{financial_phrasebank}. We have used majority voting among the annotators to determine the label invariance. 

\textbf{Filtering algorithm.}
\label{sec:filter_human}
Both the original paraphrase model and the fine-tuned model can generate up to 48 adversarial examples per original example, which are too many for a human annotator to easily evaluate. To amend this, we have filtered the set of generated adversarial examples to get a smaller representative subset that is less onerous to evaluate. 
The pseudocode for the filtering is provided in Algorithm \ref{alg:filter}.

The goal is to end up with a set of roughly 6-12 candidates per original example that approximately preserve the diversity of the larger set. In brief, the algorithm computes sentence embeddings for all candidates, applies a dimensionality reduction algorithm, UMAP \citep{UMAP}, and then applies a hierarchical clustering algorithm, HDBSCAN \citep{HDBSCAN}. 
We have eventually sampled a few candidates from each cluster, with the actual number depending on the number of clusters extracted by HDBSCAN.



\begin{algorithm}
\begin{algorithmic}

\State Select all successes $\mathbb{S}$ from candidate set $\mathbb{C} \in \mathbb{S}$
\If{ $|\mathbb{S}| \leq 6$}
    \State Return $\mathbb{S'} = \mathbb{S}$ 
\Else{}
    \State Compute sentence embeddings $E(s), \forall s \in \mathbb{S}$
    \State Reduce dimensionality to get $E'(s) , \forall s \in \mathbb{S}$ 
    \State Cluster $E'(s)$ 
    \State Create $\mathbb{S'}$ by sampling from each cluster
    \State Return $\mathbb{S'}$
\EndIf
\caption{Pseudocode for filtering the generated adversarial examples. }\label{alg:filter}
\end{algorithmic}
\end{algorithm}

\section{Examples of generated text}
\label{sec:examples}
Tables \ref{tab:examples_RT} and \ref{tab:examples_FP} show examples of the generated text. We present one example per dataset, including all the decoding methods with the best original and fine-tuned models, and the best-performing token-modification attack. For reasons of space, the examples only show eight generated adversarial candidates for each approach.

Each decoding method has exhibited specific trends. Sampling and beam search have generated examples that are very similar to each other, only differing slightly. Additionally, sampling has generated fewer unique candidates than the other methods. Both low and high-diversity beam search have generated good quality candidates, but also many that were ungrammatical or incoherent; and more so for the high-diversity case. 
Overall, we notice that the generated adversarial candidates seem to frequently contain consistent phrases or terms, which makes them \textit{universal adversarial examples} \citep{wallace-etal-2019-universal}. The training procedure clearly learns which phrases affect sentiment across many examples, and the approach inserts them into its generated solutions. This is less desirable than a model that can generate ``tailored'' paraphrases for each original example. The paraphrasing capability of the fine-tuned model is still remarkable, although at times visibly lower than that of the original paraphrase model. Overall, the fine-tuned model seems to have been able to generate better paraphrases on Financial PhraseBank than Rotten Tomatoes, likely because of its simpler language.

\begin{table*}[]
\resizebox{\textwidth}{!}{%
\small
\begin{tabular}{ll}
\hline
\textbf{Approach} &
  \multicolumn{1}{c}{\textbf{Generated Text}} \\ \hline
Original example &
  suffers from unlikable characters and a self-conscious sense of its own quirky hipness. \\ \hline
Token-modification &
  {\color[HTML]{008000} suffers from unlikable characters and gains   a self-conscious sense of its own quirky hipness .} \\ \hline
 &
  {\color[HTML]{A6A6A6} suffers from unlikable characters and has a sense of   self-conscious hipness.} \\
 &
  {\color[HTML]{A6A6A6} suffers from unlikable   characters and a self-conscious sense of its own quirky hipness.} \\
 &
  {\color[HTML]{A6A6A6} suffers from unlikable   characters and a self-conscious sense of its own quirky hipness..} \\
 &
  {\color[HTML]{A6A6A6} suffers from unlikable   characters and an in-your-face sense of quirky hipness.} \\
 &
  {\color[HTML]{A6A6A6} suffers from unlikable   characters and a self-conscious sense of its own quirky hipness...} \\
 &
  {\color[HTML]{A6A6A6} suffers from unlikable   characters and an internal sense of eccentric hipness..} \\
 &
  {\color[HTML]{A6A6A6} suffers from unlikeable   characters and a self-conscious sense of its own quirky hipness. ''} \\
\multirow{-8}{*}{Sampling, original paraphrase model} &
  {\color[HTML]{A6A6A6} suffers from unlikable   characters and self-conscious sense of humour.} \\ \hline
 &
  {\color[HTML]{A6A6A6} Unlikable characters may even have a self-conscious hipness} \\
 &
  {\color[HTML]{A6A6A6} Unlikable characters may even   have a self-conscious sense of their own quirky hip} \\
 &
  {\color[HTML]{A6A6A6} Unlikable characters may even   have self-conscious hipness} \\
 &
  {\color[HTML]{A6A6A6} Unlikable characters may even   have a self-conscious sense of their own hip} \\
 &
  {\color[HTML]{A6A6A6} Unlikable characters might   even have an self-conscious hip hop habit} \\
 &
  {\color[HTML]{A6A6A6} Unlikable characters might   even have a self-conscious hip hop attitude} \\
 &
  {\color[HTML]{A6A6A6} Unlikable characters might   even have a self-conscious sense of their own quirky hip} \\
\multirow{-8}{*}{Sampling, fine-tuned model} &
  {\color[HTML]{A6A6A6} Unlikable characters may even   have self-conscious hip-ness} \\ \hline
 &
  {\color[HTML]{A6A6A6} it suffers from unlikable characters and a self-conscious   sense of its own quirky hipness.} \\
 &
  {\color[HTML]{A6A6A6} suffers from unlikable   characters and a self-conscious sense of its own quirky hip-hop.} \\
 &
  {\color[HTML]{A6A6A6} suffers from unlikable   characters and a self-conscious sense of its own quirky hip-ness.} \\
 &
  {\color[HTML]{A6A6A6} suffers from unlikable   characters and self-conscious sense of its own quirky hipness.} \\
 &
  {\color[HTML]{A6A6A6} he suffers from unlikable   characters and a self-conscious sense of his own quirky hipness.} \\
 &
  {\color[HTML]{A6A6A6} suffers from unlikable   characters and a sense of self-consciousness of its own quirky hipness..} \\
 &
  {\color[HTML]{A6A6A6} suffers from unlikable   characters and a self-conscious sense of quirky hipness..} \\
\multirow{-8}{*}{Beam search,   original paraphrase model} &
  {\color[HTML]{A6A6A6} suffers from unremarkable   characters and a self-conscious sense of its own quirky hipness.} \\ \hline
 &
  {\color[HTML]{008000} It suffers from unlikable characters and a   self-conscious sense of its own quirky hipness but there is something} \\
 &
  {\color[HTML]{008000} It suffers from unlovable characters   and a sense of its own quirky hipness but there is something} \\
 &
  {\color[HTML]{A6A6A6} It suffers from unlikable   characters and an unconscious sense of its own quirky hipness but there is   something} \\
 &
  {\color[HTML]{008000} It suffers from unlikable characters   and an inner sense of its own quirky hipness but there is something} \\
 &
  {\color[HTML]{008000} It suffers from unlikable characters   that have a sense of their own quirky hipness but there is something} \\
 &
  {\color[HTML]{008000} It suffers from an unlikable character   and a sense of its own quirky hipness but there is something} \\
 &
  {\color[HTML]{008000} It suffers from unlikable characters   and a feeling of its own quirky hipness but there is something} \\
\multirow{-8}{*}{Beam search,   fine-tuned model} &
  {\color[HTML]{A6A6A6} It suffers from unlikable   characters and a sense of its own quirky hipness but there is some} \\ \hline
 &
  {\color[HTML]{A6A6A6} suffers from unlikable characters and a self-conscious sense   of its own quirky hipness.} \\
 &
  {\color[HTML]{A6A6A6} suffers from unlikable   characters and a self-conscious sense of its own quirky hipness. ''} \\
 &
  {\color[HTML]{A6A6A6} suffers from unlikable   characters and a self-conscious sense of its own quirky hipness.} \\
 &
  {\color[HTML]{A6A6A6} '' suffers from unlikable   characters and a self-conscious sense of its own quirky hipness} \\
 &
  {\color[HTML]{008000} a character that suffers from an   unlikable character and a self-conscious sense of its own quirky hipness} \\
 &
  {\color[HTML]{A6A6A6} there is a lack of likeable   characters and a fear of its own eccentricity.} \\
 &
  {\color[HTML]{A6A6A6} there is a lack of likeable   characters and a fear of its own quirky hips.} \\
\multirow{-8}{*}{Low-diversity beam   search, original paraphrase model} &
  {\color[HTML]{A6A6A6} there is a lack of likeable   characters and a sense of its own quirky hipness..} \\ \hline
 &
  {\color[HTML]{008000} It suffers from unlikable characters but   uses a sense of its own quirky hipness} \\
 &
  {\color[HTML]{008000} It suffers from unlikable characters   but includes a sense of its own quirky hip} \\
 &
  {\color[HTML]{008000} It suffers from unlikable characters   but often displays its own quirky hip} \\
 &
  {\color[HTML]{A6A6A6} It comes from unlikable   characters but uses a sense of its own quirky hipness} \\
 &
  {\color[HTML]{A6A6A6} It comes from unlikable   characters but includes a sense of its own quirky hipness} \\
 &
  {\color[HTML]{008000} It struggles with unlikable characters   but often displays its own quirky hip} \\
 &
  {\color[HTML]{A6A6A6} It also has unlikable   characters but is intended for its own quirky hip} \\
\multirow{-8}{*}{Low-diversity beam   search, fine-tuned model} &
  {\color[HTML]{008000} It struggles with unlikable characters   but often displays its own quirky hipness but uses some hip} \\ \hline
 &
  {\color[HTML]{008000} has unlikable characters and a sense of its   own quirky hipness.} \\
 &
  {\color[HTML]{008000} it has unlikable characters and a   self-conscious sense of its own quirky hipness.} \\
 &
  {\color[HTML]{A6A6A6} suffers from unlikable   characters and an uncanny sense of its own quirky hipness.} \\
 &
  {\color[HTML]{008000} it has unlikable characters and a sense   of its own quirky hipness.} \\
 &
  {\color[HTML]{A6A6A6} the characters are not likable   but the sense of their own quirky hipness is a problem.} \\
 &
  {\color[HTML]{A6A6A6} the characters are not likable   but the sense of their own quirky hipness is a problem.} \\
 &
  {\color[HTML]{008000} has unlikable characters and a   self-conscious sense of its own quirky hipness....} \\
\multirow{-8}{*}{High-diversity   beam search, original paraphrase model} &
  {\color[HTML]{A6A6A6} shows an unlikely character   with an unrequited feeling of hipster...} \\ \hline
 &
  {\color[HTML]{008000} The unlikable characters should have a   self-conscious sense of hipness} \\
 &
  {\color[HTML]{A6A6A6} Sometimes it should suffer   from unlikable characters but sometimes self-conscious senses of hipness} \\
 &
  {\color[HTML]{008000} Unlikable characters should have a   self-conscious sense of hipness} \\
 &
  {\color[HTML]{A6A6A6} This can involve unlikable   characters but sometimes self-conscious hipness.} \\
 &
  {\color[HTML]{008000} It must suffer from unlikeable   characters but a self-conscious sense of hipness} \\
 &
  {\color[HTML]{008000} The unlikeable characters must have a   self-conscious sense of hipness} \\
 &
  {\color[HTML]{A6A6A6} The unlikeable character   should have a self-conscious sense of its own hipness} \\
\multirow{-8}{*}{High-diversity   beam search, fine-tuned model} &
  {\color[HTML]{A6A6A6} it will have to suffer from   unlikable characters but often quirky hipness} \\ \hline
\end{tabular}%
}
\caption{Examples of generated adversarial candidates for the various approaches for an original example from the Rotten Tomatoes dataset. The original example is in black and has negative sentiment. We use light grey for unsuccessful candidates and green for a successful adversarial example that changes sentiment to positive.}
\label{tab:examples_RT}
\end{table*}

\begin{table*}[]
\large
\resizebox{.9\textwidth}{!}{%
\begin{tabular}{ll}
\hline
\textbf{Approach}                                     & \multicolumn{1}{c}{\textbf{Generated Text}}                                                                                       \\ \hline
Original example                                               & In addition , the company will reduce a maximum of ten jobs .                                              \\ \hline
Token-modification                                     & {\color[HTML]{008000} In addition , the company will reduce a   maximum requirement of ten jobs .}         \\ \hline
                                                       & {\color[HTML]{A6A6A6} In addition, the company can reduce the jobs of up to ten   people.}                 \\
                                                       & {\color[HTML]{A6A6A6} In addition the company will   cut a maximum of ten jobs.}                           \\
                                                       & {\color[HTML]{1E90FF} In addition he says the company will   close a maximum of ten jobs.}                 \\
                                                       & {\color[HTML]{A6A6A6} In addition, the company will   eliminate up to ten jobs.}                           \\
                                                       & {\color[HTML]{A6A6A6} In addition the company will   reduce up to ten jobs.}                               \\
                                                       & {\color[HTML]{A6A6A6} In addition the company has   cut a maximum of ten jobs.}                            \\
                                                       & {\color[HTML]{A6A6A6} In addition '' the company   will cut 10 jobs in 10 months.}                         \\
\multirow{-8}{*}{Sampling, original paraphrase model}                    & {\color[HTML]{A6A6A6} In addition to the job cuts   the company will limit itself to ten jobs.}            \\ \hline
                                                       & {\color[HTML]{A6A6A6} In addition, the company has succeeded in reducing a maximum   of ten jobs.}         \\
                                                       & {\color[HTML]{A6A6A6} In addition, the company has   succeeded in decreasing a maximum of ten jobs.}       \\
                                                       & {\color[HTML]{008000} In addition, the company has succeeded   in cutting a maximum of ten jobs.}          \\
\multirow{-4}{*}{Sampling, fine-tuned model}                      & {\color[HTML]{A6A6A6} In addition, the company has   succeeded inreducing a maximum of ten jobs.}          \\ \hline
                                                       & {\color[HTML]{A6A6A6} In addition, the company will reduce a maximum of ten jobs.}                         \\
                                                       & {\color[HTML]{A6A6A6} In addition he will reduce a   maximum of ten jobs.}                                 \\
                                                       & {\color[HTML]{A6A6A6} In addition, the company will   reduce a maximum of ten jobs.}                       \\
                                                       & {\color[HTML]{A6A6A6} The company will also reduce a   maximum of ten jobs.}                               \\
                                                       & {\color[HTML]{A6A6A6} In addition, the company will   cut a maximum of ten jobs. ''}                       \\
                                                       & {\color[HTML]{A6A6A6} In addition the company will   cut up to ten jobs.}                                  \\
                                                       & {\color[HTML]{A6A6A6} In addition, the company will   eliminate a maximum of ten jobs.}                    \\
\multirow{-8}{*}{Beam search,   original paraphrase model}             & {\color[HTML]{A6A6A6} In addition the company will   cut a maximum of ten jobs..}                          \\ \hline
                                                       & {\color[HTML]{A6A6A6} In addition, the company will also be successful in reducing a   maximum of 10 jobs} \\
                                                       & {\color[HTML]{008000} In addition, the company will also see   success in reducing a maximum of 10 jobs}   \\
                                                       & {\color[HTML]{008000} In addition, the company will also see   success in reducing a maximum of 10 jobs.}  \\
                                                       & {\color[HTML]{008000} In addition, the company will also see   success in reducing ten jobs.}              \\
                                                       & {\color[HTML]{008000} In addition, the company will also see   success at reducing a maximum of 10 jobs.}  \\
                                                       & {\color[HTML]{A6A6A6} In addition, the company will   also have success in cutting a maximum of 10 jobs.}  \\
                                                       & {\color[HTML]{008000} In addition, the company will also find   success in reducing a maximum of 10 jobs}  \\
\multirow{-8}{*}{Beam search,   fine-tuned model}                  & {\color[HTML]{008000} In addition, the company will also find   success in reducing a maximum of 10 jobs.} \\ \hline
                                                       & {\color[HTML]{A6A6A6} In addition, the company will reduce a maximum of ten jobs.}                         \\
                                                       & {\color[HTML]{A6A6A6} In addition, the company will   reduce a maximum of ten jobs..}                      \\
                                                       & {\color[HTML]{A6A6A6} In addition, the company will   reduce a maximum of ten jobs. ''}                    \\
                                                       & {\color[HTML]{A6A6A6} In addition the company will   cut a maximum of ten jobs. ''}                        \\
                                                       & {\color[HTML]{A6A6A6} In addition, the company will   reduce a maximum of ten jobs...}                     \\
                                                       & {\color[HTML]{A6A6A6} In addition, the company will   reduce a maximum of ten jobs in the region.}         \\
                                                       & {\color[HTML]{A6A6A6} He said: In addition he would   reduce the company's total number of jobs to 10.}    \\
\multirow{-8}{*}{Low-diversity beam   search, original paraphrase model}  & {\color[HTML]{1E90FF} In addition, the company will reduce a   maximum of ten job cuts.}                   \\ \hline
                                                       & {\color[HTML]{A6A6A6} In addition, the company will benefit from reducere ten jobs.}                       \\
                                                       & {\color[HTML]{A6A6A6} This is good for decreasing a   maximum of 10 jobs.}                                 \\
                                                       & {\color[HTML]{A6A6A6} Along with this reduction of   10 jobs, the company will benefit from reductions.}   \\
                                                       & {\color[HTML]{A6A6A6} The company will benefit from   r$\sqrt{}$©duiing a maximum of 10-jobs.}                     \\
                                                       & {\color[HTML]{008000} In addition, the company will benefit   from reductions for ten jobs.}               \\
                                                       & {\color[HTML]{008000} In addition the company will benefit   from reductions for the maximum of 10 jobs.}  \\
                                                       & {\color[HTML]{008000} As a result the company will benefit   from reductions of up to 10 jobs.}            \\
\multirow{-8}{*}{Low-diversity beam   search, fine-tuned model} & {\color[HTML]{008000} The company will benefit from reducing   ten jobs.}                                  \\ \hline
                                                       & {\color[HTML]{A6A6A6} The company will also reduce a maximum of ten jobs.}                                 \\
                                                       & {\color[HTML]{A6A6A6} A further reduction of ten   jobs is planned. ''}                                    \\
                                                       & {\color[HTML]{A6A6A6} A maximum of ten jobs will be   lost in the company.}                                \\
                                                       & {\color[HTML]{A6A6A6} Furthermore, the company will   reduce up to ten jobs..}                             \\
                                                       & {\color[HTML]{A6A6A6} The company will reduce its   workforce by up to 10 percent. ''}                     \\
                                                       & {\color[HTML]{A6A6A6} It also plans to eliminate at   least ten jobs in the company..}                     \\
                                                       & {\color[HTML]{A6A6A6} Further the company will   reduce a maximum of 10 jobs.}                             \\
\multirow{-8}{*}{High-diversity   beam search, original paraphrase model} & {\color[HTML]{A6A6A6} Additionally the company will   lose up to ten jobs.}                                \\ \hline
                                                       & {\color[HTML]{1E90FF} The company also reduced the reduction of a   maximum of ten jobs.}                  \\
                                                       & {\color[HTML]{008000} Furthermore the company will reduce the   reduction of ten jobs.}                    \\
                                                       & {\color[HTML]{A6A6A6} This also improved reduction   of up to ten jobs.}                                   \\
                                                       & {\color[HTML]{1E90FF} The company also reduced the reduction   of ten jobs..}                              \\
                                                       & {\color[HTML]{008000} the company will also improve   reductions for up to ten jobs.}                      \\
                                                       & {\color[HTML]{A6A6A6} Also the company would reduce   a maximum of ten jobs.}                              \\
                                                       & {\color[HTML]{1E90FF} The company also reduced the reduction   of ten jobs...}                             \\
\multirow{-8}{*}{High-diversity   beam search, fine-tuned model}   & {\color[HTML]{008000} Additionally the firm would also   improve the reduction of the maximum of 10 jobs.} \\ \hline
\end{tabular}%
}
\caption{Examples of generated adversarial candidates for the various approaches for an original example from the Financial PhraseBank dataset. The original example is in black and has negative sentiment. We use light grey for unsuccessful candidates, and green and blue for successful adversarial examples that changed the sentiment to positive and neutral, respectively. (NB: the ``Sampling, fine-tuned model'' approach only produced four unique candidates for this example.) }

\label{tab:examples_FP}
\end{table*}

\end{document}

%% file: table_ablation_examples.tex
\begin{table}[h]
\centering
\footnotesize
\begin{tabularx}{\columnwidth}{l>{\raggedright\arraybackslash}Xl}
\toprule
\textbf{Method} & \textbf{Text} & \textbf{Label} \\
\midrule
Original & Where does one find Rider College? & Loc \\
LM-based, low-budget (BAE-R) & or can one find Rider College? & Enty \\
LM-based, high-budget (variant 2)  & So does one find Rider College? & Desc \\
Fine-tuned model & How does one find Rider College? & Abbr \\
\midrule
Original & How can I get some free technical information on Electric Vehicle? & Desc \\
LM-based, low-budget (BAE-R) & When can I get some basic technical information on Electric Vehicle? & Num  \\
LM-based, high-budget (variant 2)  & When can I get some free technical information on Electric Vehicle? & Num  \\
Fine-tuned model & can someone help me get free technical information about electric vehicles? & Enty \\ \midrule
Original & What do the Japanese call Japan? & Enty \\
LM-based, low-budget (BAE-R) & What do the Japanese mean Japan? & Desc \\
LM-based, high-budget (variant 2)  & What do the Japanese in Japan? & Desc \\
Fine-tuned model & How do the Japanese call Japan? & Abbr \\    
\bottomrule
\end{tabularx}
\caption{Examples of successful attacks generated by the various approaches from samples from the TREC dataset (NB: the untrained paraphrase model did not produce any successful attack from these samples). Note that the compared approaches have somehow distorted the original meaning with their token transformations, while our fine-tuned model has been able to retain it more closely.}
\label{tab:ablation_examples}
\end{table}